\documentclass[conference]{IEEEtran}
\IEEEoverridecommandlockouts
\usepackage{amsmath,amssymb,amsfonts}
\usepackage{algorithmic}
\usepackage{graphicx}
\usepackage{textcomp}
\usepackage{xcolor}
\usepackage{multirow}
\usepackage{booktabs}
\usepackage{subfigure}
\usepackage[numbers]{natbib}
\def\BibTeX{{\rm B\kern-.05em{\sc i\kern-.025em b}\kern-.08em
    T\kern-.1667em\lower.7ex\hbox{E}\kern-.125emX}}
\begin{document}

\title{Faint Features Tell: Automatic Vertebrae Fracture Screening Assisted by Contrastive Learning
\thanks{This work was supported in part by National Natural Science Foundation of China (62106248) and Medical Health Science and Technology Project of Zhejiang Provincial Health Commission (2022KY1188)

*Equal Contribution.

\dag Corresponding Authors: Guoping Chen(headoniones@aliyun.com) and Jinpeng Li(lijinpeng@ucas.ac.cn)}}

\DeclareRobustCommand*{\IEEEauthorrefmark}[1]{%
  \raisebox{0pt}[0pt][0pt]{\textsuperscript{\footnotesize #1}}%
}
\author{
    \IEEEauthorblockN{
        \textbf{Xin Wei*}\IEEEauthorrefmark{1,2}, 
        \textbf{Huaiwei Cong*}\IEEEauthorrefmark{1,2}, 
        \textbf{Zheng Zhang}\IEEEauthorrefmark{3}, 
        \textbf{Junran Peng}\IEEEauthorrefmark{4}, 
        \textbf{Guoping Chen\dag}\IEEEauthorrefmark{1} and
        \textbf{Jinpeng Li\dag}\IEEEauthorrefmark{1,2}
    }\\
    \IEEEauthorblockA{
        \IEEEauthorrefmark{1}HwaMei Hospital, University of Chinese Academy of Sciences (UCAS), Ningbo, China\\
        \IEEEauthorrefmark{2}Ningbo Institute of Life and Health Industry, UCAS, Ningbo, China\\
        \IEEEauthorrefmark{3}Department of Computer Science and Technology, Harbin Institute of Technology, Shenzhen\\
        \IEEEauthorrefmark{4}National Laboratory of Pattern Recognition, Institute of Automation, Chinese Academy of Sciences\\
    }
}

\maketitle

\begin{abstract}
Long-term vertebral fractures severely affect the life quality of patients, causing kyphotic, lumbar deformity and even paralysis. Computed tomography (CT) is a common clinical examination to screen for this disease at early stages. However, the faint radiological appearances and unspecific symptoms lead to a high risk of missed diagnosis, especially for the mild vertebral fractures. In this paper, we argue that reinforcing the faint fracture features to encourage the inter-class separability is the key to improving the accuracy. Motivated by this, we propose a supervised contrastive learning based model to estimate Genent's Grade of vertebral fracture with CT scans. The supervised contrastive learning, as an auxiliary task, narrows the distance of features within the same class while pushing others away, enhancing the model's capability of capturing subtle features of vertebral fractures. Our method has a specificity of 99\% and a sensitivity of 85\% in binary classification, and a macro-F1 of 77\% in multi-class classification, indicating that contrastive learning significantly improves the accuracy of vertebrae fracture screening. Considering the lack of datasets in this field, we construct a database including 208 samples annotated by experienced radiologists. Our desensitized data and codes will be made publicly available for the community.
\end{abstract}

\begin{IEEEkeywords}
deep learning, vertebral fracture, contrastive learning, computer-aided diagnosis
\end{IEEEkeywords}

\begin{figure*}[htbp]
\centerline{\includegraphics[width=0.95\textwidth]{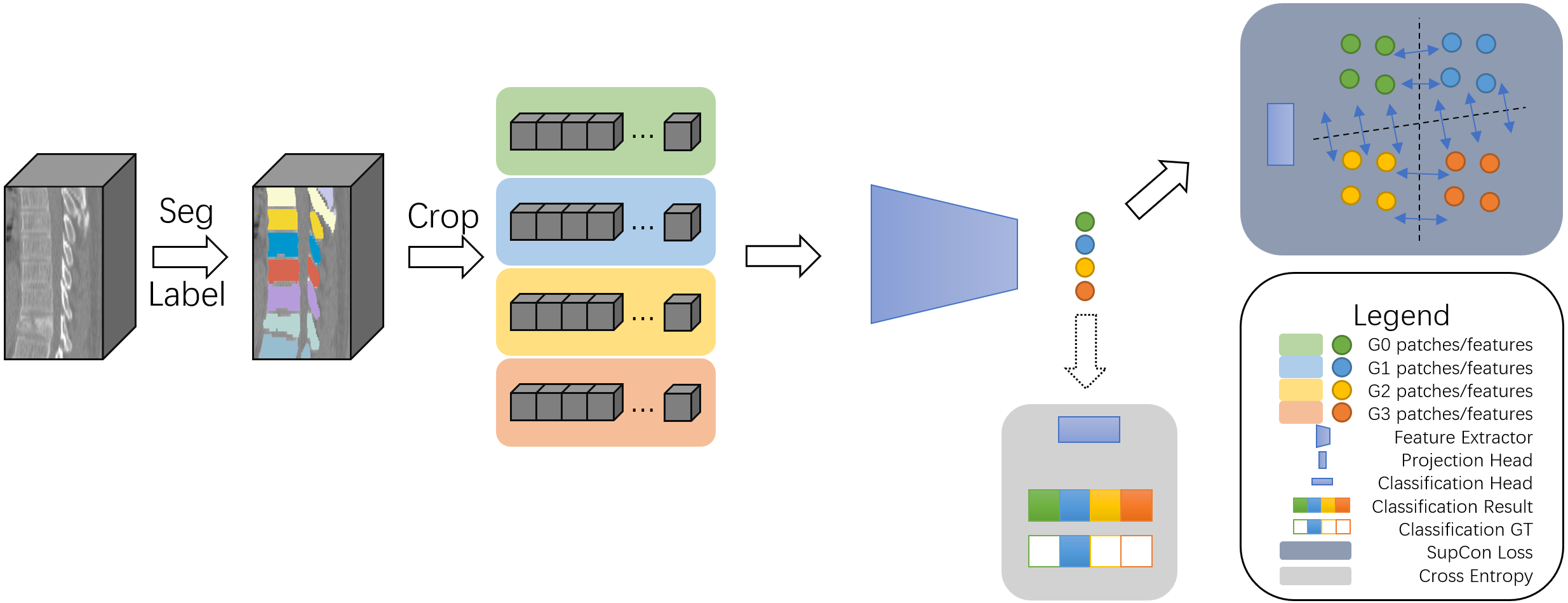}}
\caption{Overview of our pipeline. Our pipeline is arranged in a two-stage manner. The vertebrae in the CT scans are segmented and labeled first to get the patches of vertebrae. It followed by the grading network which consists of a feature extractor, a projection head and a classification head. The features of patches are calculated by the feature extractor, and further clustered by fracture grade with the projection head via supervised contrastive learning. The grading results are given by the classification head, which won't propagate gradient to the feature extractor (marked as dotted arrow in the figure).}
\label{overview}
\end{figure*}

\section{Introduction}
Vertebral fracture is a common disease that often occurs in aged people, which could severely affect patients' living. Deformity and chronic pain are the major clinical manifestations of vertebral fracture, and according to \citet{risk}, long-term vertebral fractures can increase the mortality rate by eight times. However, such dangerous disease is often under-reported in clinical diagnosis. This is due to its less specific symptoms and less obvious radiological appearances, leading to radiologists' neglect or misattribution\cite{missdiag}. Since early diagnosis and treatment are essential for alleviating vertebral fractures' impact on human health, a computer-aided screening tool with high performance could be fairly helpful.

Currently, Genant semi-quantitative method\cite{genant} serves as the common standard in assessing vertebral fractures, but it will be affected by radiologists' experience and awareness of scrutinizing vertebrae radiography. On the other hand, deep learning based methods could be automatically inferred and thus excluding radiologists' subjectivity, makes it an ideal method to screen vertebral fracture. Due to the slight difference among classes, the grading of vertebrae can be regarded as a fine-grained classification task. To address this, we propose a supervised contrastive learning based method to enhance the inconsistent feature among each grade. We validated our method on a dataset collected by us and a public dataset, both of them show improvements on classification by a large margin. 


Our contribution can be summarized as follows:
\begin{itemize}
\item We design an end-to-end pipeline to segment, label and assess fracture on each vertebra of the given CT image. Such assessment procedure is fully automatic and without any human intervention, which excludes influence of radiologists' subjectivity. We believe such approach could largely improve the under-diagnosed situation of vertebral fracture.
\item We propose to utilize supervised contrastive learning in vertebrae fracture grading, and design series of studies to prove that our method has a better capability of detecting mild vertebral fracture. It can be concluded that forming feature space by contrastive learning further drives CNN to capture the information in the given images. 
\item To validate our method, we collected and arranged a novel vertebrae dataset that contains spine CT images of various fracture situations. Our dataset is collected from real clinical cases, which are well aligned and have suitable resolution for analysis. To support the research community of medical image analysis, \textbf{we will publicly share our desensitized dataset} shortly, together with the codes of this paper. 
\end{itemize}

\section{Related Works}

\subsection{Vertebrae Segmentation and Labeling}
Segmentation and labeling of vertebrae are the fundamental tasks for further processing and analysis, for reliable segmentation and labeling algorithm could enable multiple automatic assessing tasks such as vertebral fracture grading or spine deformity detection. For this reason, vertebrae segmentation and labeling keep drawing research communities' attention. In 2019 and 2020, The Large Scale Vertebrae Segmentation Challenge (Verse) was held in conjunction with MICCAI, evaluating multiple algorithms of vertebrae segmentation and labeling. We highly suggest readers who are interesting in this topic referring to the report of the challenges\cite{versereport}. In this paper, we utilize the algorithm of \citet{payer}, which introduces a U-net\cite{unet} based structure and segments vertebrae in a coarse-to-fine manner.

\subsection{Vertebrae Fracture Grading}
On the other hand, studies about vertebrae fractures grading are relatively insufficient. Unlike segmentation and labeling tasks, grading of fractures is facing naturally imbalanced data for abnormal vertebrae only account for a small portion of overall vertebrae. This makes an enlarged data demand for fracture grading tasks. Since vertebral fracture is direct related to the deformity of vertebrae, conventional methods segment vertebrae to be assessed in the CT images, and calculate its shape statistic with the segmentation mask\cite{cseg1,cseg2,cseg3}. However, severe deformities like burst fractures could degrade the segmentation algorithm, limiting its major application to osteoporosis and compression fractures. \citet{cresnet} assesses vertebrae fracture with neighboring CT slices, and \citet{clstm} aggregates feature in CT slices with a LSTM\cite{lstm}. \citet{cplain} also evaluates deep learning model's performance of vertebral fracture detection on plain spinal radiography. Similar to our idea, \citet{c3d} managed to detect vertebral fractures in CT volumes with a 3D CNN, and referring to metric learning, \citet{cgrading} proposed a novel metric loss that could form a reasonable feature space. In this paper, we further demonstrate that supervised contrastive learning could reinforce the faint feature and form a better clustered feature space, resulting in advanced performance for vertebral fracture grading.

\subsection{Contrastive Learning}
Contrastive learning aims at forming a clustered feature space to enhance feature extraction. Major contrastive learning methods focus on self-supervised learning, for clustering could be achieved only by appearances of given images, removing the necessity of manual annotation. The clustering is often made by narrowing the distance among positive samples while enlarging that among negative samples. In self-supervised scenario, the positive samples are two distinct views from the same item, while the negative samples are views form the others. MoCo\cite{moco} designed a memory mechanism to expand the negative samples form mini-batch to a dynamic memory bank, and SimCLR\cite{simclr} carefully researched major factors of contrastive learning, finding the importance of projection head, data augmentation and batchsize. After that, they shared the idea mutually and came up with the updated version SimCLRv2\cite{simclrv2} and MoCov2\cite{mocov2}. 

Contrastive learning could also improve fully-supervised learning. With fully annotated label, SupCon\cite{supcon} expanding positive samples to same-class samples, showing a better performance as well as a more robust optimization comparing to cross entropy loss. In this paper, we follow the idea of SupCon\cite{supcon} and further demonstrate its capability of fine-grained classification on medical images.


\section{Method}



\subsection{Dataset}

Generally, deep learning is a data-driven approach which requires massive annotated data to converge. However, annotating vertebral fracture of CT images is difficult and error-prone. To overcome this issue, we use a deep learning method to aid the annotation procedure, alleviating the workload as well as improving the quality of the annotation. Specifically, we first adopt \citet{payer} to segment and label vertebrae in the CT volume. It could segment and label vertebrae with a Dice coefficient at 0.93, which is suitable for our application. By utilizing this automatic segmentation approach, radiologists could avoid manual segmentation of each vertebra in the CT scans. In practice, \citet{payer} could generally give precise segmentation masks and labels. Its major mistake is the occasionally mislabeling, which will be revised by our invited radiologists later.

Another issue of annotation is the faint radiological appearances of vertebral fractures. This may lead to inconsistent annotation, which brings ambiguous feature that dramatically hurt the clustering of contrastive learning. To address this, we first invited 3 junior radiologists to assess fractures of the vertebrae we segmented. The radiologists are entrusted to annotate each vertebra with its Ganent's Grade, as well as revise the segmentation or label if the algorithm\cite{payer} gives incorrect output. For the disagreements in annotation, we invited a senior radiologist to verify the voting of initial annotation and give the conclusion. With the proposed annotation, we could arrange a high-quality vertebral fracture dataset with rather light workload.

Our method does not restrict to certain population, so we make no assumption about gender or age of participants we collected in the dataset. The generalization capability of deep learning model could promise the model's applicability for broad population. We balanced the ratio of each Ganent's Grade in the adopted participants to relieve potential data imbalance. Genant's Grade classifies the fractures into 4 classes of G0, G1, G2 and G3, which can be regarded as normal, mild, moderate and severe. In practice, we adopt 208 CT scans in our dataset with the ratio of G0:G1:G2:G3=1:3:3:3, including 2,423 vertebrae in total.






\subsection{Model}
Our CNN architect is designed with a two-stage manner. Vertebrae in the CT volume are cropped to patches first with an existing segmentation and labeling model. Then the vertebrae patches will be fill into the grading network to estimate the Ganent's Grade of each vertebra with a supervised contrastive learning manner. An overview of our pipeline is illustrated in Fig.~\ref{overview}, and the detailed design is stated as follows:

\emph{Segmentation and Labeling}
We adopt \citet{payer} as the segmentation and labeling model, which we also utilized in dataset annotation. The output segmentation masks are expanded to bounding box first to include nearby tissue as an additional clue for fracture grading, then the patches of vertebrae are cropped accordingly. we choose two windows to extract information from original CT patches: the bone window that sets window level with 1500 HU and window width with 400 HU, and the soft tissue window that sets window level with 200 HU and window width with 40 HU. To our knowledge, such windowing contains sufficient information for vertebrae fracture grading.

We also take the original segmentation mask and label into account, by concatenating the segmentation mask to the input patches of CNN. Label information is also helpful, for underlying fracture-related feature of different vertebrae are potentially varied. We modulate the original binary segmentation mask with its normalized label and concatenate it with the two windows of CT image at channel dimension. The combined three channel image will be used as input of the subsequent grading network, with an additional resampling to ensure the isotropic voxel spacing and uniform orientation.

\emph{Network Structure}
We follow the contrastive learning methods\cite{simclr,supcon} to design our grading network, which can be separated into three parts. First, a backbone network acts as the \textbf{feature extractor} to extracts and encodes the feature of radiograph. Then a \textbf{projection head} projects the feature to a low-dimensional space, where the optimization of contrastive learning applied. Additionally, a \textbf{classification head} takes the output of feature extractor and estimates the grading result, with a cross entropy loss to optimize. In practice, shallow networks like several linear layers could fulfill the intention of projection head and classification head. Also, as contrastive learning methods have proven that the feature space is separable without any information of classification head, we follow the advice of SupCon\cite{supcon} to detach the classification head so that its gradient won't be back propagated to other part of the network. At inference time, the projection head will be discarded, leaving output of classification head as the grading result. A graphic demonstration of our network is illustrated in Fig.~\ref{overview}.

We adopt 3D-SEnet50\cite{senet} as the structure of feature extractor, which adds an attention mechanism to the conventional ResNet50\cite{resnet} model. It shows a better feature extracting capability especially for the 3D fine-grained classification. The projection head and the classification head both consist of a single linear layer, with a 128-dimensional output for the projection head, and a 4-dimensional output for the classification head of 4-level grading. We also follow the convention to normalize the magnitude of the 128-dimensional vector to 1 for a spherical space is better for contrastive learning.

\begin{table*}[htbp]
\caption{Quantitative Result. FE, SPE, SEN are short for Feature Extractor, Specificity and Sensitivity.}
\begin{center}
\begin{tabular}{cccccccccc}
\toprule  
\multirow{2}{*}{FE}&\multirow{2}{*}{Loss}&\multirow{2}{*}{Optimizer}&\multirow{2}{*}{Dataset}&\multicolumn{3}{c}{Binary Classification}&\multicolumn{3}{c}{Multi-Class Classification}\\
\cmidrule(lr){5-7}\cmidrule(lr){8-10}
&&&&AUCROC&SPE&SEN&Macro-F1&Macro-Precision&Macro-Recall\\
\midrule  
ResNet50&Cross Entropy&Adam&Ours&0.95&0.96&0.72&0.53&0.52&0.56\\
ResNet50&SupCon&SGD&Ours&0.98&0.99&0.83&0.67&0.67&0.65\\
SENet50&Cross Entropy&Adam&Ours&0.97&0.99&0.79&0.59&0.61&0.57\\
SENet50&SupCon&Adam&Ours&0.98&0.97&\textbf{0.87}&0.64&0.62&0.67\\
\textbf{SENet50}&\textbf{SupCon}&\textbf{SGD}&Ours&\textbf{0.98}&\textbf{0.99}&0.85&\textbf{0.77}&\textbf{0.78}&\textbf{0.77}\\
\midrule
SENet50&Cross Entropy&Adam&Verse&0.90&0.94&0.68&0.59&0.59&0.62\\
\textbf{SENet50}&\textbf{SupCon}&\textbf{SGD}&Verse&\textbf{0.93}&\textbf{0.94}&\textbf{0.72}&\textbf{0.72}&\textbf{0.72}&\textbf{0.72}\\
\midrule
\citet{cgrading}&Grading Loss&Adam&Verse w/o G1&-&0.99&0.77&-&-&-\\
\textbf{SENet50}&\textbf{SupCon}&\textbf{SGD}&Verse w/o G1&\textbf{0.99}&\textbf{0.99}&\textbf{0.88}&\textbf{0.86}&\textbf{0.84}&\textbf{0.85}\\
\bottomrule 
\end{tabular}
\label{abla}
\end{center}
\end{table*}

\emph{Loss Function}
Contrastive learning methods augment input sample to a pair of distinct views. For the self-supervised manner, the clustering is conducted by taking the pair of views as positive sample mutually while the others as negative samples. This manner can cluster appearance-similar features without additional labels, makes it a feasible and prevalent solution for self-supervised learning.

On the other hand, supervised contrastive learning keeps utilizing annotated labels to guide the clustering of feature space, with the main assumption that samples in the same class could vary in appearance. We keep using supervised manner for the faint disparities among fracture grades is much weaker than disparities in the overall appearance of vertebrae. \citet{fgfail} also gives an observation that fine-grained classification could degrade the self-supervised contrastive learning. However, we argue that with the guidance of class label, the contrastive learning method is encouraged to detect the fine-grained disparities which contribute most to the task.

We adopt SupCon loss\cite{supcon} as the loss function of our method, with the equation in \eqref{supcon}
\begin{equation}
  \mathcal{L}^{sup}
  =\sum_{i}\frac{-1}{|P(i)|}\sum_{p\in P(i)}\log{\frac{\exp\left(\text{sim}\left(\mathbf{z}_{i}, \mathbf{z}_{p}\right)/\tau\right)}{\sum_{k}\mathbb{I}_{[k\neq{i}]}\exp\left(\text{sim}\left(\mathbf{z}_{i}, \mathbf{z}_{k}\right)/\tau\right)}}
  \label{supcon}
\end{equation}

It is a enhanced version of NT-Xent loss\cite{simclr} which expanded the positive sample set $P(i)$ to same-class samples, i.e. $P(i)\equiv\{p|\boldsymbol{\tilde{y}}_p=\boldsymbol{\tilde{y}}_i,p\neq i\}$. Here the $\mathbf{z}$ is the feature vector of projection head, and the temperature parameter $\tau$ is used to control the intensity of loss. With SupCon loss, the disparities among classes become the major clues of classification, and in the radiographic diagnosis scenario, such disparities strongly hint regions of lesion. To prove this, we use Grad-CAM\cite{gradcam} to visualize some results to check whether features of lesion region attract high attention. The visualization can be found in the \textit{Results} section.


\emph{Data augmentation}
For contrastive learning, data augmentation is used to generate multiple distinct views of original training data. SimCLR\cite{simclr} carefully researched the combinations of different data augmentations and their impact on the contrastive learning. However, as \citet{databias} mentioned, the detailed configuration of data augmentation is often data-biased, especially for our task that is distinct from classification of natural images. Intuitively, data augmentation should try to avoid interfering the clue of classification, but regular data augmentation cannot promise this, for the vertebral fracture are related to global feature like posture and shape, as well as the local feature like minor fractures. To address this, as listed below, we design a set of data augmentations that are specialized for vertebral fracture grading:

\begin{itemize}
\item Random Padding.
\item Pre-Rotation Mask of 2 boxes with side lengths of 1-20 voxels.
\item Random Zoom in 0.9-1.1x.
\item Gaussian Noise with $\mu=0$ and $\sigma=0.05$.
\item Random Shift in $\pm$10 voxels.
\item Random Rotation in $\pm$10 degrees.
\item CT Value Jittering with the function $$HU=(HU \times p1)^{p2}$$
, where p1 is within 0.9-1.1 and p2 is within $\pm$2.
\item Post-Rotation Mask of 2 boxes with side lengths of 1-20 voxels.
\end{itemize}

Random Padding takes an input patch and rescale its longest side length to 128, and further randomly padding it to the resolution of 3*128*128*128(3 are the channel dimension introduced in section \emph{Segmentation and Labeling}). The other data augmentations are applied with possibilities of 70\%. Multi-dimensional augmentations are applied to each axis with individual parameters. The result shows that it reaches a good trade-off between distinct views and fracture clue reserving. 

As introduced in \emph{Loss Function} section, input vertebra patches will be augmented twice individually to generate the pair of views. The pair will be accumulated to a mini-batch and fed into the subsequent grading network.

\subsection{Training}
We specially designed several training techniques to facilitate the contrastive learning. In this section, we will introduce the details in the training procedure.

\emph{Batch Sampler}
Ordinary batch sampler often traverses the dataset randomly in an epoch. However, since vertebral fracture is happened occasionally, its fracture grade is naturally facing data imbalance. And as for the supervised contrastive learning, this issue is non-negligible for the clusters may be optimized unevenly, causing a biased grading. To avoid this, we design a Per-Class Sampler, which randomly sample $n$ patches in each class, forming a mini-batch of $nC$ patches where $C$ is the number of classes. The sampling procedure is without placement, and it resets every time when the least class traversed. Less rigorously, we still call it an 'epoch', and giving enough epochs, the dataset could be traversed with a data-adaptive sampling rate. This is a simple yet powerful batch sampler, which could avoid data imbalance as well as accelerate the converge of optimization.


\emph{Optimizer}
We adopt SGD as the optimizer with a weight decay of 1e-4 and a momentum of 0.9. For the supervised contrastive learning, it converges to a better minimum than adaptive optimizers like Adam\cite{adam} in practical. The learning rate starts at 1e-3, and decays by 0.1 at epoch 800 and 900.

\emph{Miscellaneous}
We implement our methods with Pytorch\cite{pytorch} on a workstation with two RTX A6000 graphic cards. The dataset is split into training set and test set with the ratio of 4:1, and vertebrae of individuals won't be split into different set. We sample 6 vertebrae patches per class at each iteration, forming a batchsize of 24 in total. Each patch is augmented twice to generate the pair of views. Due to the data limitation, the batchsize we set is relatively small comparing to that on natural images, but the result is still impressive, showing the potential of contrastive learning on medical images. On our workstation, the network takes about 18 hours to converge, with approximately 1000 epochs.

\begin{figure}[htbp]
\centering

\subfigure[Attention Map, generated with Grad-CAM\cite{gradcam}]{
\begin{minipage}[t]{0.95\linewidth}
\centering
\includegraphics[width=1in]{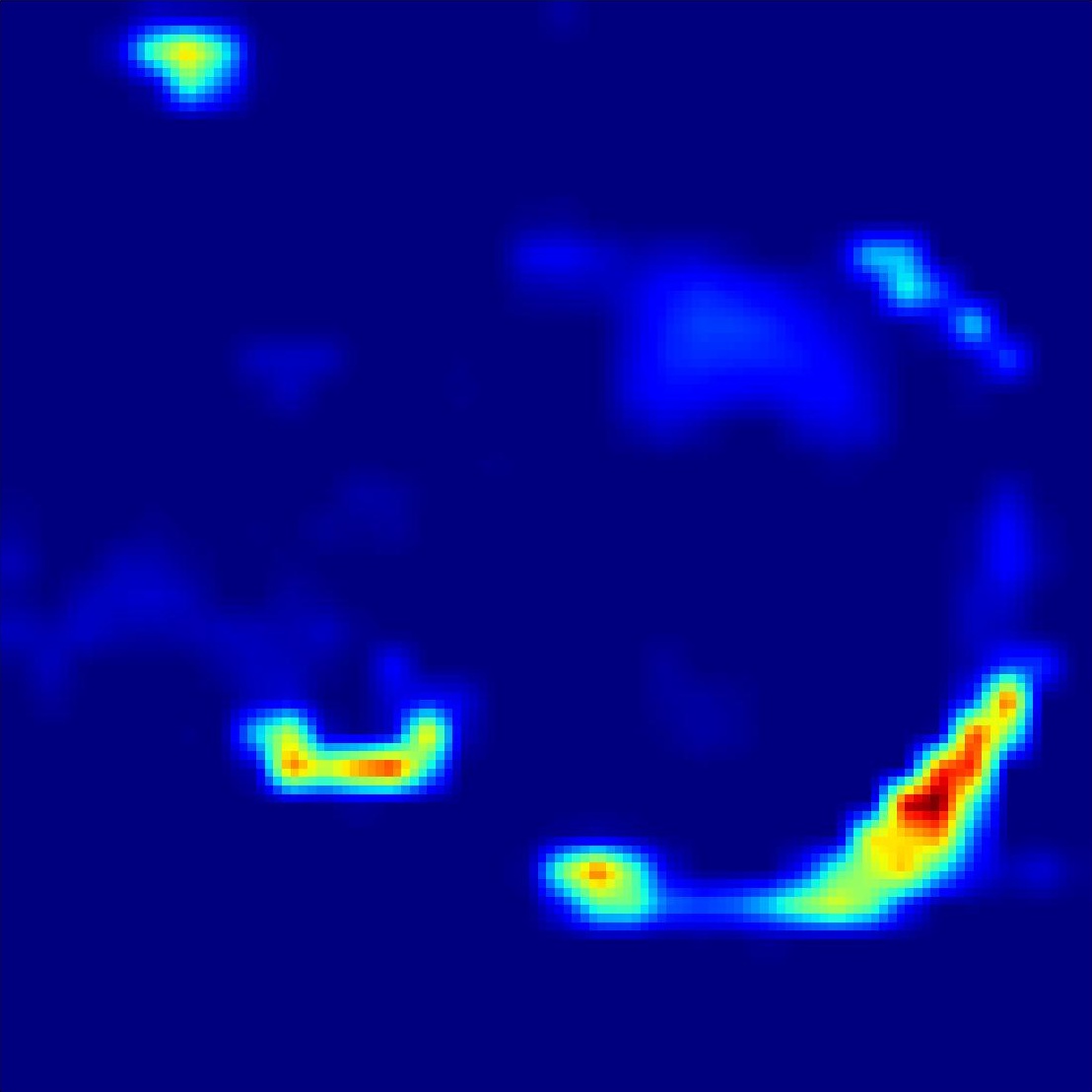}
\includegraphics[width=1in]{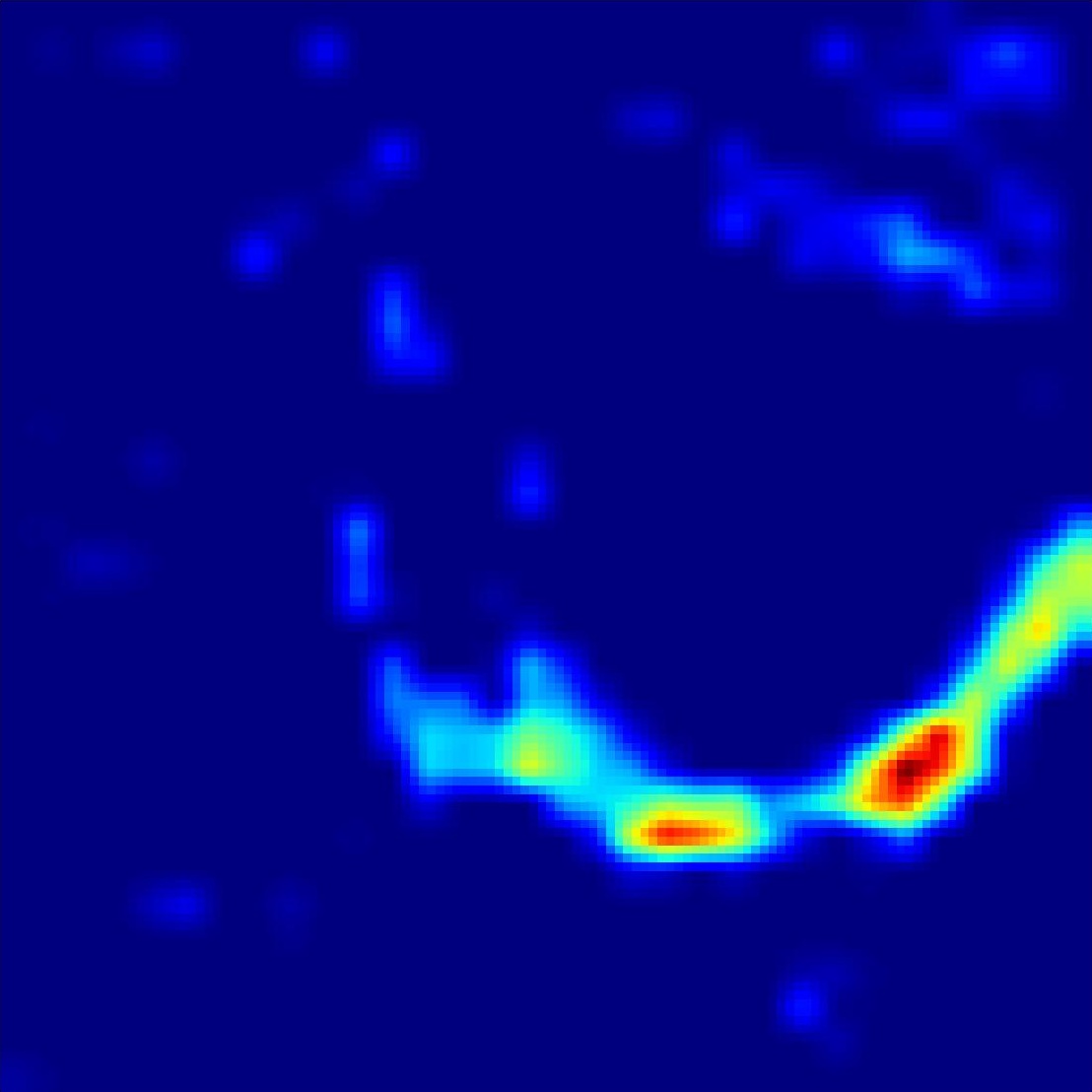}
\includegraphics[width=1in]{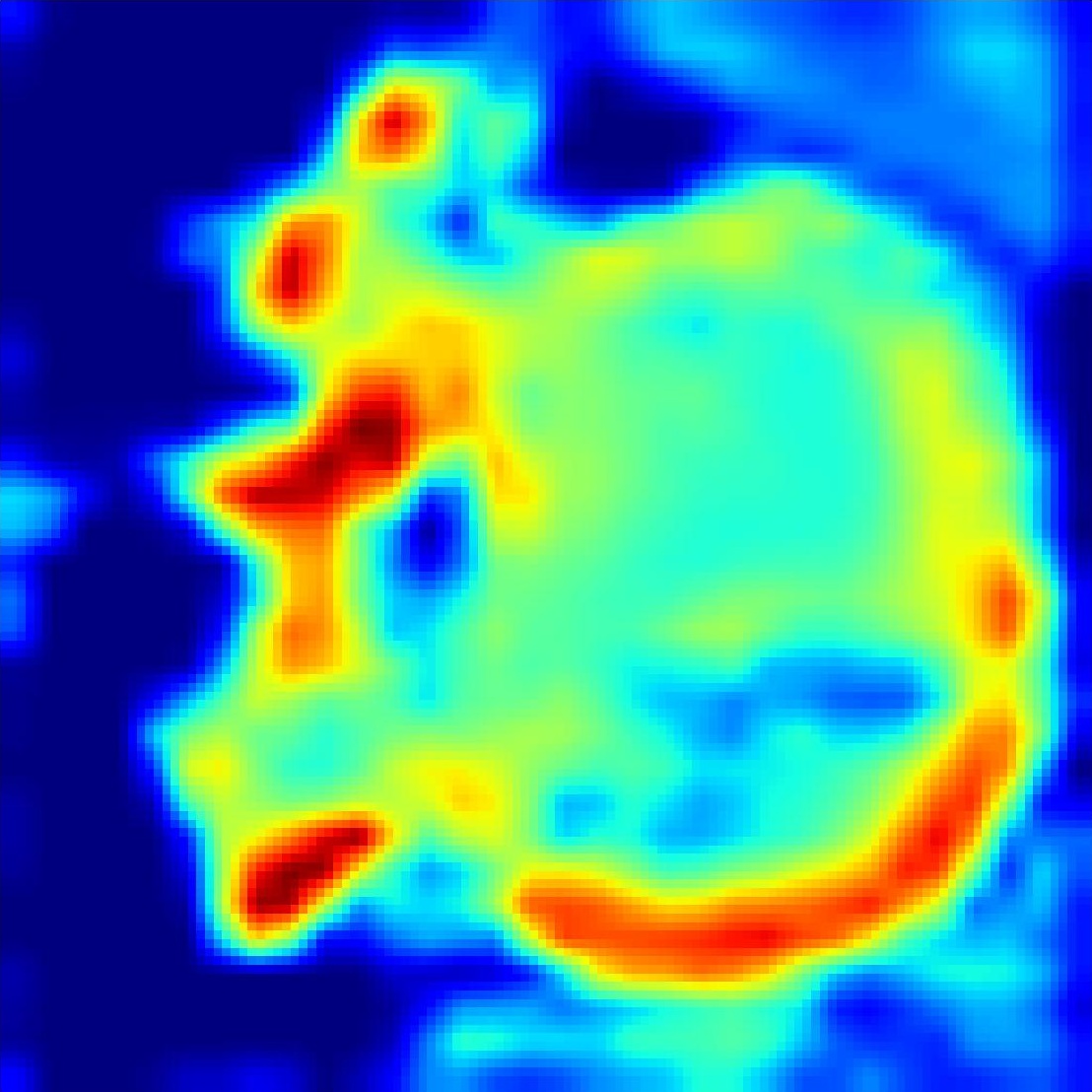}
\end{minipage}%
}%

\subfigure[Corresponding CT Slices]{
\begin{minipage}[t]{0.95\linewidth}
\centering
\includegraphics[width=1in]{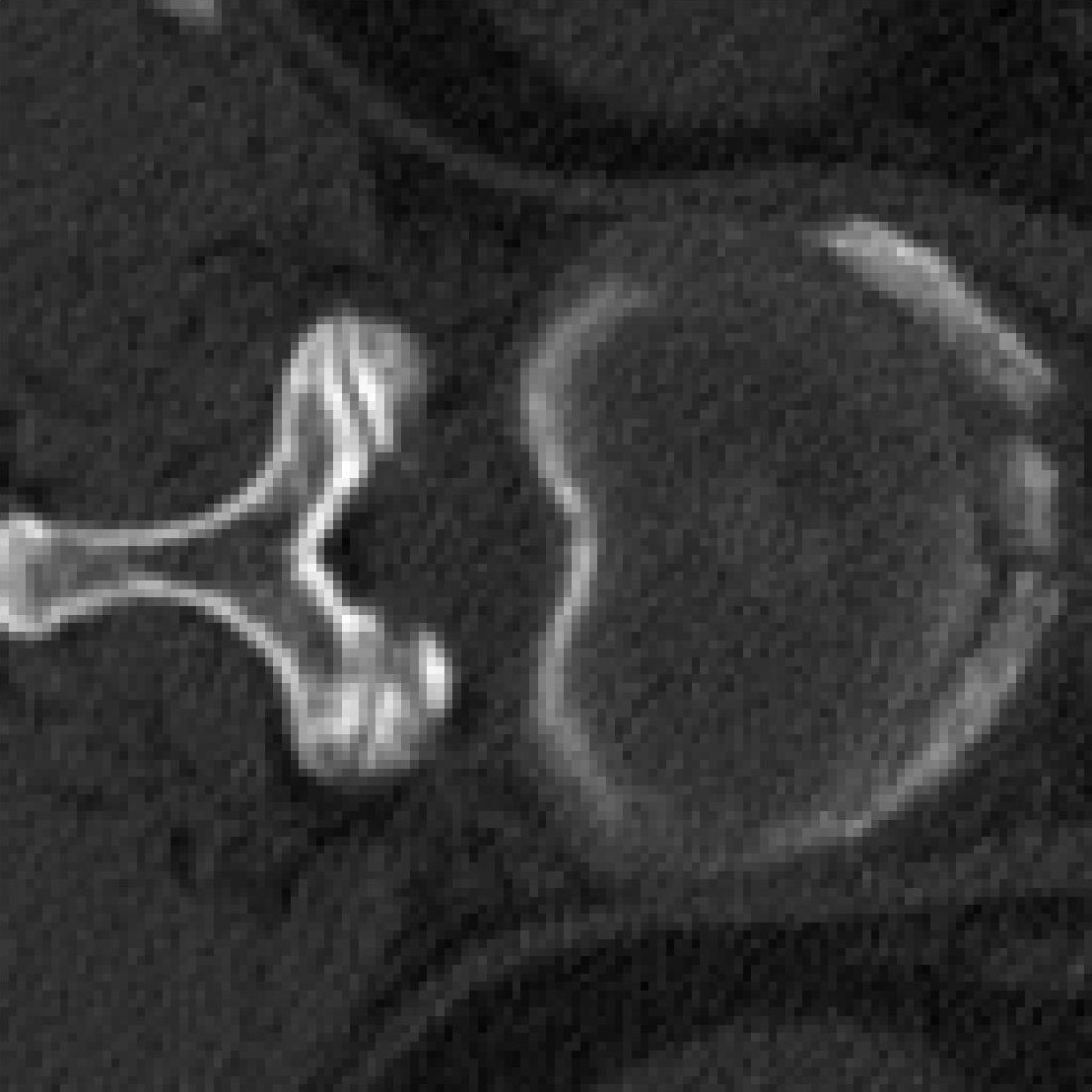}
\includegraphics[width=1in]{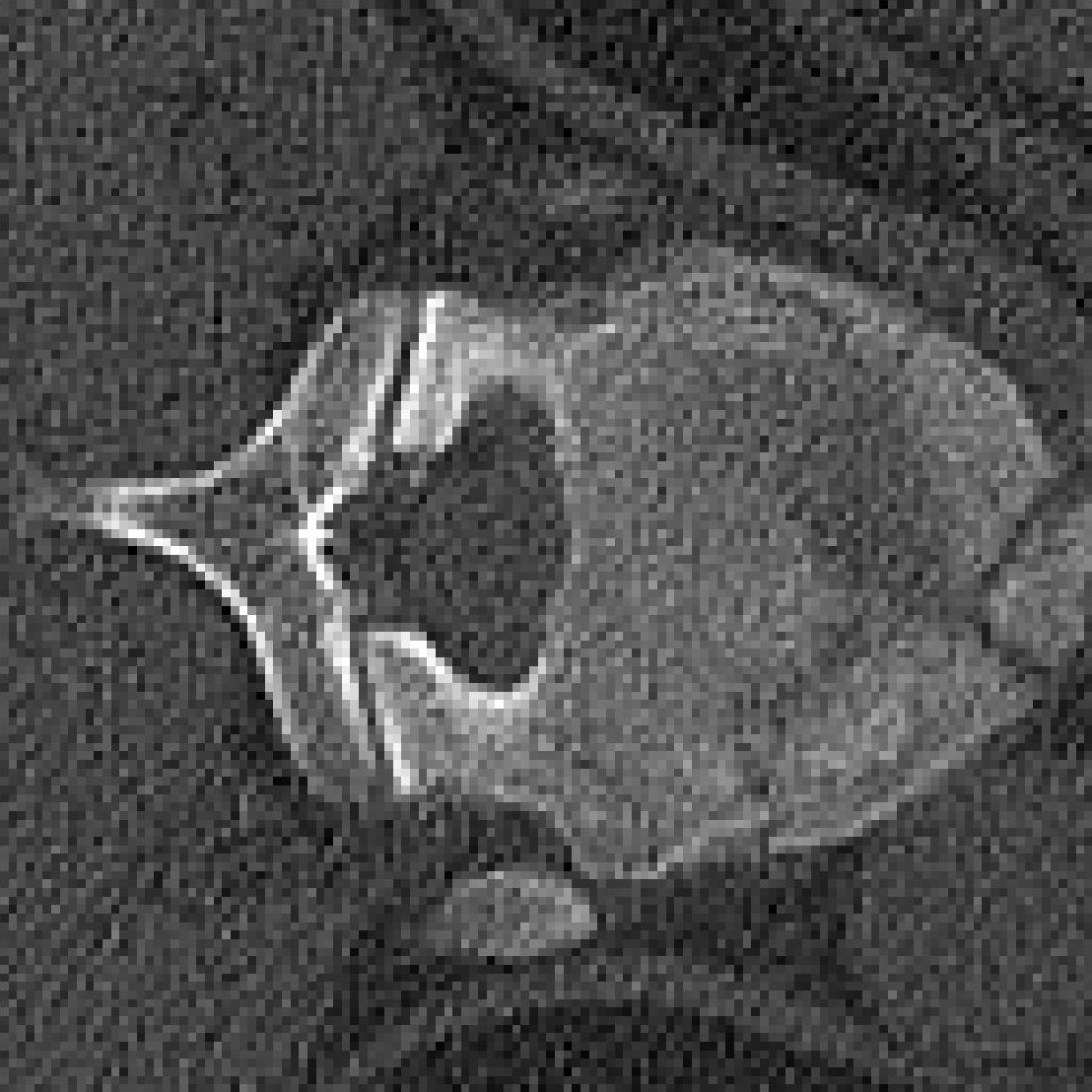}
\includegraphics[width=1in]{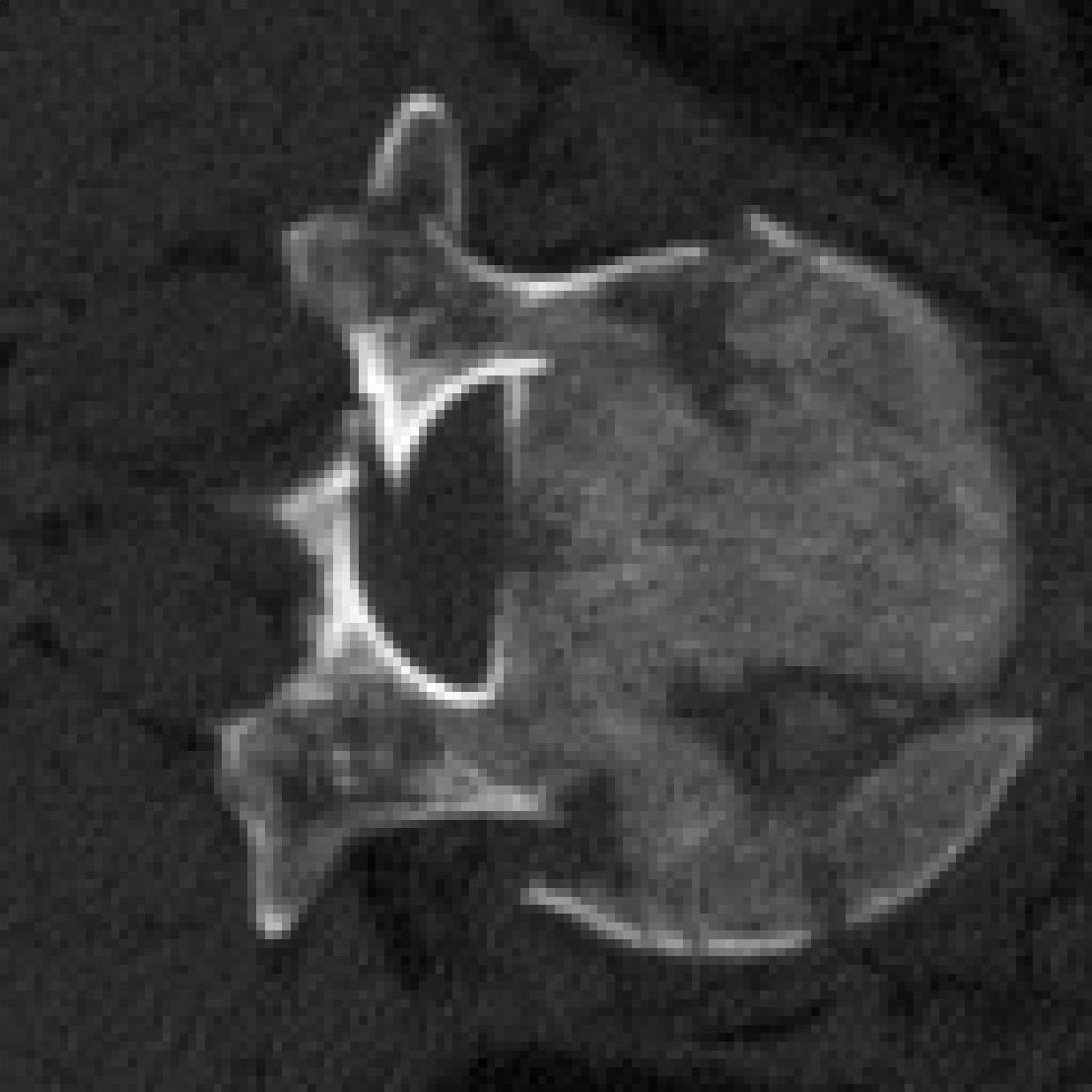}
\end{minipage}%
}%


\centering
\caption{Qualitative Result. The Genant's Grade of the CT slices are G1, G2 and G3 respectively.}
\label{cam}
\end{figure}

\section{Results}

\subsection{Quantitative Study}

\emph{Metrics}
We mainly evaluate two aspects of our method, which are the \textbf{binary classification} of Benign(G0) vs Malignant(combination of G1,G2 and G3), as well as the \textbf{multi-class classification} of 4 grades. The intention of evaluating binary classification is that it's the most sensitive metrics for missed diagnosis.

To avoid the misleading of unbalanced test set, we use AUCROC as the main metrics of binary classification, and macro-F1 score as the main metrics of multi-class classification. Note that we always take multi-class classification as the training target, while the evaluation of binary classification is only happened at inference time by combining the class G1, G2 and G3 in ground truth and prediction respectively.

\begin{figure}[htbp]
\centerline{\includegraphics[width=0.8\linewidth]{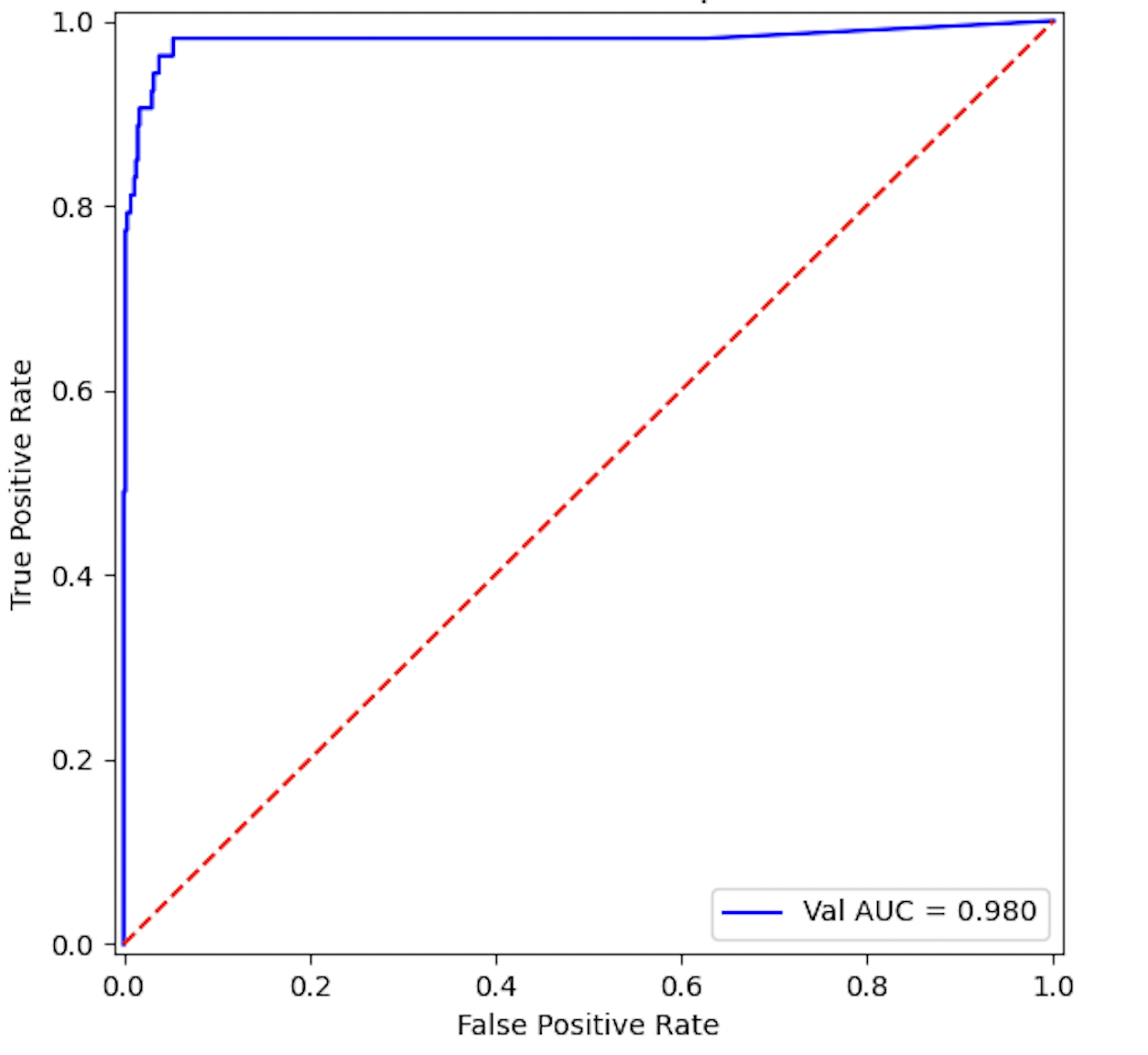}}
\caption{ The ROC curve of binary classification.}
\label{roc}
\end{figure}

\emph{ROC curve}
Firstly, we demonstrate the binary classification performance of our method with the ROC curve in Fig.~\ref{roc}. With a specificity of 99\% and sensitivity of 85\%, it could improve the diagnostic rate of vertebral fracture by a large margin.

\emph{Ablation study}
To prove the aforementioned benefits we claimed about our methods, we evaluate the basic ResNet50 models with cross entropy loss, and gradually add it to the final version. The detailed experiments and results are listed in Table.~\ref{abla}, and our full model as well as the best result are marked in bold. Note that SupCon loss largely improve the multi-class classification result comparing to cross entropy, which is the key contribution of our work.

\emph{Public Dataset Validation}
To avoid being data biased, we also evaluate our method on a public vertebral fracture dataset Verse\cite{versereport}. As a challenge dataset, it contains more complicated situation than our clinical dataset, e.g., more varied resolution and orientation, as well as additional cervical vertebrae. These situations cause degraded metrics than our dataset, however the improvement of supervised contrastive learning is still remarkable. Also, our dataset does not contain any vertebrae with artificial implants, for they actually do not need screening. To keep a reasonable comparison, we remove vertebrae with artificial implants in Verse\cite{versedata}. The result can be found in Table.~\ref{abla}.

\emph{Comparative study}
We choose \citet{cgrading} to conduct the comparative study, for it was validated on Verse\cite{versedata} as well. With the auxiliary information from fracture grade, it managed to improve the binary classification with a novel grading loss. Due to the difficulty in distinguishing Grade 0 and Grade 1, \citet{cgrading} didn't take Grade 1 fractures into account, while ours could detect the mild Grade 1 fractures and further enable multi-class classification. For reference, we also validate our method on Verse\cite{versedata} without Grade 1 fractures. The result can be found in Table.~\ref{abla}.

\subsection{Qualitative study}
As we mentioned, supervised contrastive learning picks the feature that strongly hints the region of lesion, and Grad-CAM\cite{gradcam} is a proper tool to visualize such region. The volumetric Grad-CAM\cite{gradcam} is generated with the implement of \citet{cam_impl}. As show in Fig.~\ref{cam}, with our method, the model could detect the regions of vertebral fractures in multiple situations.

\section{Conclusion}

We design a pipeline of vertebral fracture grading with supervised contrastive learning, which shows a great performance in both binary and multi-class classification. We believe our method could improve the diagnostic rate of vertebral fracture in real clinical scenario. Also, we arranged a high-quality vertebral fracture dataset with careful annotations of Genant's Grade, which may alleviate the data deficiency of related research.


\small
\bibliographystyle{IEEEtranN}
\bibliography{citation}  
\end{document}